\documentclass{article}
\usepackage[utf8]{inputenc}
\usepackage[T1]{fontenc}
\usepackage{url}
\usepackage{booktabs}
\usepackage{amsfonts}
\usepackage{amsmath}
\usepackage{nicefrac}
\usepackage{microtype}
\usepackage{xcolor}
\usepackage{graphicx}
\usepackage{multirow}
\usepackage{lipsum}
\usepackage{comment}
\usepackage{amsmath,amssymb,amsthm,booktabs}
\theoremstyle{plain}

\theoremstyle{definition}

\theoremstyle{remark}

\PassOptionsToPackage{numbers, compress, sort}{natbib}
\usepackage[dblblindworkshop, final]{neurips_2026}

\title{Critical Mass: Structure-Disjoint Splits Drift With the Libraries They Are Built On}
\workshoptitle{AI Data Readiness for Scientific Discovery}

\author{%
  Tomo Oga\thanks{Corresponding author: \texttt{tomo@matterworks.ai}} \quad Devesh Shah \quad Cailum M. K. Stienstra \quad Gabriel Asher \\
  \textbf{Antonio H. de O. Fonseca \quad Niall O'Connor \quad Michael Widrich}\\ [0.3em]
  Matterworks, Inc.\\
  Somerville, MA 02143, USA\\
}

\begin{document}
\maketitle

\begin{abstract}
    Molecular machine learning evaluates out-of-distribution performance with structure disjoint splits, a construction rule fixed once at the scale its library happened to have: threshold a pairwise distance, then assign whole connected components to folds. For predicting structures from mass spectra, the field has converged on single-linkage clustering under the myopic Maximum Common Edge Subgraph (MCES) distance at a threshold of 10. This convention was set on 29,000 molecules but public libraries now hold 227,690 molecules. MCES is NP-hard per pair over a quadratically growing set, and so the criterion has never been evaluated at that scale. We propose an optimized version of this criterion, allowing for computationally feasible application on the public libraries: Admissible bounds and a union-find skip reduce $2.6 \times 10^{10}$ molecular pairs to only $1.15 \times 10^{7}$ integer programs, a $2{,}259\times$ saving, while provably producing the lossless partition. We find that at this scale the graph percolates. A single component holds $95.5\%$ of the library, capping any test set at $4.5\%$. What remains is not merely smaller, but systematically different: a held-out compound outweighs a training compound 88.1\% of the time, because a fixed absolute budget is a tighter relative tolerance on a larger molecule. Large library benchmarks using this splitting rule naively therefore conflate structural novelty with extrapolation to larger, more complex chemistry.
\end{abstract}

\section{Introduction}
Shared benchmarks have driven much of the progress in machine learning, though how well a reported result transfers depends on the split it was built on \cite{koh2021wilds, sogaard2021we}. The problem is acute in molecular machine learning, where random partitions place near-identical structures in different splits and reward recall of the training data over generalization \cite{wallach2018most}. Structure-disjoint evaluation has therefore become central to establishing whether a model generalizes beyond known chemical neighborhoods \cite{bushuiev2024massspecgym, liu2026wild}. Such splits compute pairwise distances between molecules, retain only the edges below a threshold $T$, and assign each connected component to a single fold. Which distance should define \textit{near}, however, remains an open question \cite{moleculenet, butina, kretschmer2025coverage, bushuiev2024massspecgym}.

In MS/MS structure elucidation, deep learning based methods have been developed to translate spectra to structure \textit{de novo} or retrieve them from candidate sets \cite{duhrkop2019sirius, goldman2023mist, bohde2025diffms, bohde2026frigid, neo2025molforge}, and are evaluated out-of-distribution on MassSpecGym \cite{bushuiev2024massspecgym}. 
This benchmark adopts the myopic Maximum Common Edge Subgraph (MCES) distance, an absolute edit distance on the molecular graph, and requires every test compound to lie at distance $>10$ (MCES-10) from every training compound, enforced by single-linkage clustering of the \textit{near} graph, arguing that MCES is more interpretable and reflects biochemical modification more faithfully than faster alternatives such as fingerprint Tanimoto.

However, fixed-threshold single-linkage partitions are sensitive to sampling density, so the same rule applied to a grown library may define a fundamentally different benchmark. Each molecule can form edges that can bridge previously separate regions, until percolation yields a single giant component \cite{penrose2003rgg,callaway2000network}. The effect is documented in chemical similarity networks \cite{amoroso2023making} and sequence-identity benchmarks \cite{petti2022constructing}, where larger datasets yield fewer separable clusters. Percolation can occur for single-linkage splits under any distance; what an \textit{absolute} threshold determines is which compounds survive it, since a fixed budget of $T$ edits perturbs a small molecule far more than a large one. This is no longer a hypothetical concern. Public ML libraries for MS/MS have since grown from $\sim$29,000 structures to 227,690 unique structures, following the release of SpectraVerse \cite{spectraverse2025} and Enveda-180 \cite{enveda180}, contributing 44,240 and 183,476 respectively. At this scale MCES-10 is also computationally intractable: MCES is NP-hard, and molecular pairs grow quadratically with library size \cite{kretschmer2025coverage}.

\textbf{Contributions.}
We show that MCES-10 partitioning defines a benchmark that is not invariant to library size. Its separation guarantee holds at every scale, but the holdout it permits collapses. Moreover, the compounds that remain are systematically heavier, because an absolute threshold is a size-dependent relative one. To measure this, we introduce an optimized MCES that makes splitting feasible at modern library sizes, reducing $2.6\times10^{10}$ pairs to $1.15\times10^7$ integer programs while preserving the exact partition.
\section{Methods}
\begin{figure}[t]
    \centering
    \includegraphics[width=\textwidth]{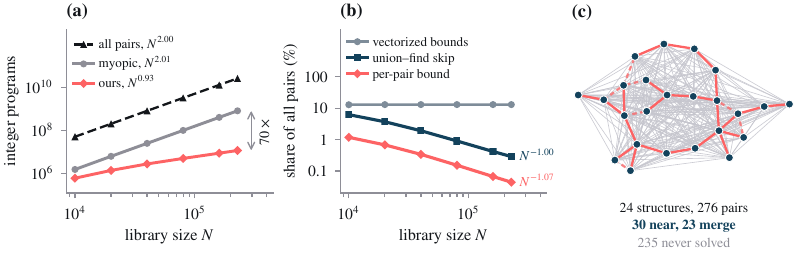}
    \caption{\textbf{The union-find skip turns a quadratic problem into a near-linear one.} (a) Integer programs solved against library size $N$: all pairs and myopic-MCES both scale as $N^2$, ours as $N^{0.93}$, a  $70\times$ gap at the full union. (b) The union-skip is the only stage that improves substantially with $N$. (c) One 24-structure cluster: of the 276 pairs among its members, 30 were solved near, 23 merging components, and 7 closing cycles, 11 were solved and returned far, and 235 never reached the solver. Myopic-MCES counts in (a) are calculated with $5\times10^6$-pair samples per rung (Appendix~\ref{app:validation}).}
    \label{fig:funnel}
\end{figure}

\textbf{Problem formulation.} MCES distance $d(A, B)$ between structures $A$ and $B$ uses a molecular graph to represent each heavy atom as a vertex carrying an element label and each bond as an edge carrying a weight $w(e)$ \cite{kretschmer2025coverage}.
It then minimizes, over all one-to-one pairings of same-element atoms, the total weight of unpaired bonds plus the weight difference of the paired ones (Appendix~\ref{app:distance}).

Clusters are then constructed such that $\{A, B\}$ are within a component (\textit{near}) when $d(A, B) < T$, and \emph{far} otherwise.
To create MCES-10 splits MassSpecGym used
myopic MCES \cite{kretschmer2025coverage}, which computes admissible bounds,
such that only pairs for which the bound cannot place beyond $T$ reach the integer program.
Myopic MCES introduces two such bounds:
\texttt{filter1} sorts both molecules' atoms by weighted degree and compares the two sequences per position,
since atoms whose connectivity differs must be reconciled by bond edits;
\texttt{filter2} optimally matches atoms of the same element by the similarity of their local neighborhoods, giving a strictly tighter bound at cubic cost.

\textbf{Exploiting the partition.}\label{sec:exploit_the_partition}
We propose two algorithmic modifications to enable the computation of myopic MCES at larger scale: \textit{Expanded bounds} and \textit{union-find skipping}.
In \textit{expanded bounds},
we expand the admissible bounds of \citet{kretschmer2025coverage} by
(a) the difference in total bond weight $|W(A)-W(B)|$,
since any excess weight in one molecule has no partner in the other,
and (b) the $l_1$ distance between bond-weight histograms of unordered pairs of endpoint elements,
since a surplus of one bond class cannot match outside it.
Both are applied vectorized and ascending in cost alongside \texttt{filter1} (Appendix~\ref{app:bounds}).
In \textit{union-find skipping},
we leverage Kruskal's cycle rule \cite{kruskal1956shortest}:
a pair whose endpoints are already connected by proven near pairs cannot change the partition, whatever its verdict.
We bucket candidates by ascending bound and solve cheapest first over a union-find,
skipping any pair whose endpoints already share a component.
Because every bound is admissible,
no near pair is ever discarded,
Kruskal's cycle rule further guarantees that skipping a pair whose endpoints are already joined by proven near edges cannot change the components.
Pruning is therefore lossless and returns the partition that solving all $\binom{N}{2}$ pairs with the same solver would (Appendix~\ref{app:union_skip}).
The solver runs under a five seconds budget. Pairs that exhaust it are conservatively treated as near, which prevents leakage (Appendix~\ref{app:bracket}).

\textbf{Experimental design.}
\label{sec:experimental-design}
Public MS/MS reference data is concentrated in SpectraVerse 1.0.1 \cite{spectraverse2025} and Enveda-180 \cite{enveda180},
contributing 44,240 and 183,476 structures respectively.
Applying SpectraVerse's standardization pipeline \cite{spectraverse2025}
gives a public union of 227,690 unique canonical SMILES.
As the two libraries overlap in only 26 structures and cover different mass ranges (Figure~\ref{fig:mass_split}a),
any mass bias in a split of the union could reflect composition rather than the criterion.
Thus, we evaluate both the union and Enveda-180,
where Enveda-180 provides the largest amount of spectra and is narrow enough in mass range to discourage formation of mass-based holdouts.
We subsample each arm at $\{10, 20, 40, 80, 160\} \times 10^3$ unique compounds and the full libraries,
using three (first three rungs) and two (otherwise) random seeds (Appendix \ref{app:rung}).
Each subsample is split by the protocol in Section \ref{sec:exploit_the_partition} with threshold $T=10$.

We report two metrics to measure how MCES-10 splits change with scale. The \textit{achievable holdout} $1-g/N$, where $g$ is the largest cluster, upper-bounds the fraction of the library available for evaluation, since the giant cluster is indivisible. The \textit{mass AUC} is the probability that a uniformly drawn holdout compound is heavier than a uniformly drawn training compound, where 0.5 indicates no bias.

\section{Results \& Discussion}\label{results}

\textbf{MCES integer program reduction.}
The public union induces $2.6\times10^{10}$ pairs, $43\times$ more than MassSpecGym, and materializes a 207GB dense distance matrix for clustering alone. The vectorized bounds certify 87\% of pairs as \textit{far}, followed by a 97.8\% reduction by union-find skip (Figure~\ref{fig:funnel}), and \texttt{filter2} applied per pair inside the solver to $1.15\times10^7$ integer programs (0.04\% of all pairs), resulting in a $2,259\times$ reduction (Appendix~\ref{app:implementation-details}). 

\begin{figure}[t]
    \centering
    \includegraphics[width=\textwidth]{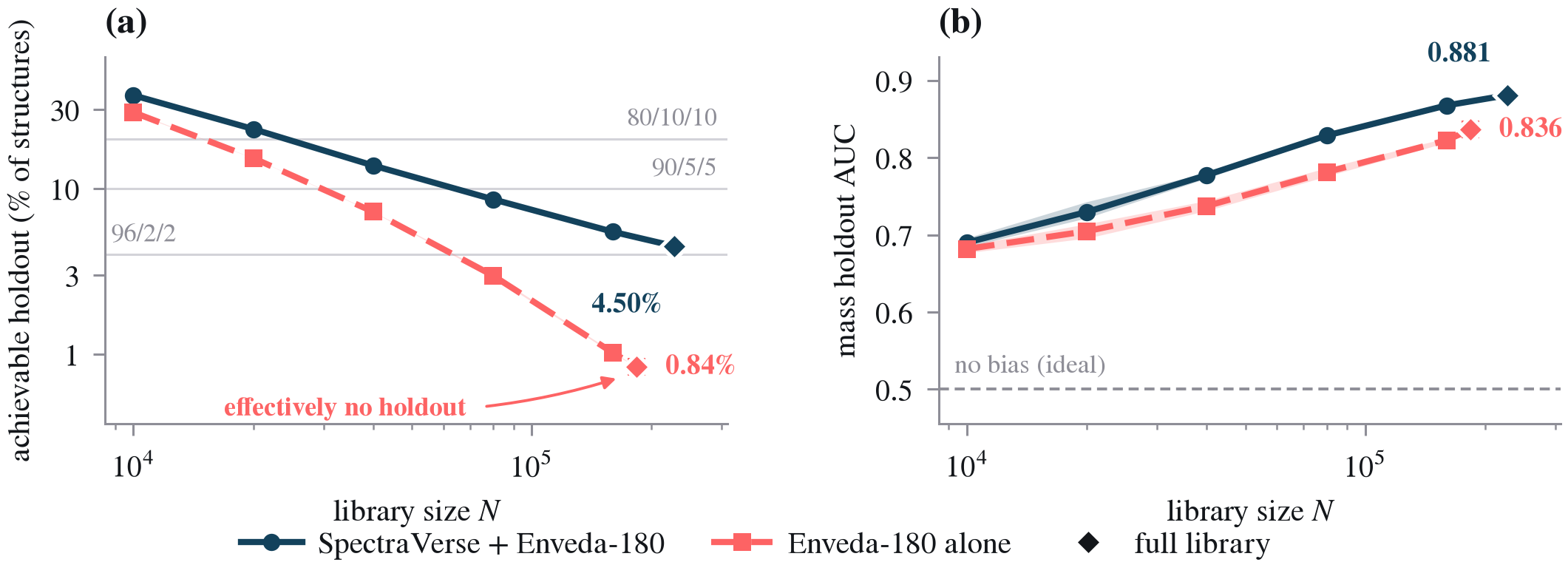}
    \caption{\textbf{The vanishing holdout of MCES with increasing dataset size skews high mass}. Structure-disjoint splits at MCES-10. (a) The largest cluster is indivisible, so the fraction outside it, $1-g/N$, upper-bounds any evaluation set; horizontal lines mark what each split ratio would require. (b) The mass holdout AUC, the probability that a compound outside the giant component is heavier than one inside it, reaches 0.881 for the union and 0.836 for Enveda-180. Bands are min-max across seeds.}
    \label{fig:scale_n}
\end{figure}

\textbf{The achievable holdout collapses, and its mass bias grows with it.}
The holdout fraction for generalization estimation under an MCES-10 graph split shrinks with increasing library size (Figure~\ref{fig:scale_n}a), falling to 4.5\% for the union and 0.84\% for Enveda-180. Moreover, the probability that a holdout compound is heavier than a training compound increases monotonically, from $\approx0.70$ ROC AUC at $N=10^4$ to $\approx0.88$ at $N=10^5$ (Figure~\ref{fig:scale_n}b). The MCES holdout therefore becomes not only smaller, but progressively enriched in high-mass structures as libraries scale.

\begin{figure}[t]
    \centering
    \includegraphics[width=\linewidth]{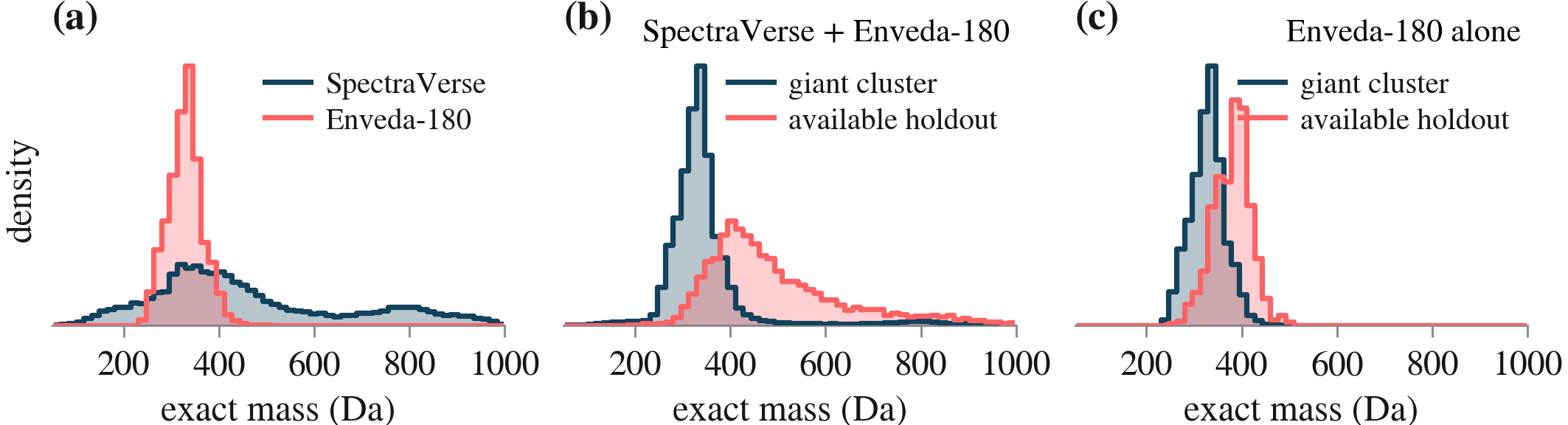}
    \caption{\textbf{MCES splitting results in high-mass skew for holdout set, irrespective of dataset mass distributions.} (a) The two libraries the public union pools. Enveda-180 supplies 80.6\% of the union and its shape dominates. (b) The union's giant component against the available holdout. (c) The same on Enveda-180 alone. }
    \label{fig:mass_split}
\end{figure}

\textbf{The bias is a property of MCES, not the library.}
The two arms have markedly different mass profiles: Enveda-180 dominates the shape of the union distribution (Figure~\ref{fig:mass_split}a), while SpectraVerse contributes substantially more high-mass compounds, making the union more representative of the broader experimental MS/MS landscape. Conventional MCES splitting nonetheless skews the holdout toward high mass in both arms (Figure~\ref{fig:mass_split}b,c). The effect is strongest for the union (Figure~\ref{fig:mass_split}b), whose holdout carries a long tail above 500~Da largely absent from training. The bias persists even for Enveda-180, where the underlying distribution is far more narrow (IQR 44~Da) and has little room for a mass split to occur, with median mass 329.2~Da in training against 384.3~Da in holdout.

\textbf{An absolute threshold is a relative one.}
Under single-linkage clustering, a molecule joins the giant component as soon as one connection falls below threshold. Ten edge edits is a large fraction of a small molecular graph but a small fraction of a large one. Low-mass molecules cross that threshold more readily and denser sampling at larger $N$ gives them progressively more opportunities to do so (Appendix~\ref{app:band}). Because the distance is unnormalized, larger molecules stay comparatively unconnected under the same absolute threshold and survive in the small components left for holdout, concentrating the split in the high-mass regime as library size grows.

\textbf{Consequences for \textit{de novo} structure elucidation.}
Whether these observed effects achieve the aim of MCES-10 splits as envisioned by the original authors is doubtful \cite{bushuiev2024massspecgym, kretschmer2025coverage}. Holdout compounds remain structurally distinct, but that distinction now arrives partly through molecular mass. Because fragmentation complexity grows with molecular size, the benchmark conflates structural out-of-distribution generalization with extrapolation across mass and fragmentation regime, a difficulty imposed by the splitting procedure rather than by structural novelty.

\textbf{Limitations.}
Both experimental libraries used in this study contain meaningful biases in their compound-mass distributions. These biases are complementary (e.g., Enveda is relatively narrow and skewed toward lighter compounds) but benchmarking on more regular mass distributions would be preferable. However, our goal is to evaluate MCES under realistic deployment conditions for MS/MS prediction, which requires reflecting the experimental libraries currently available. Public MS/MS datasets have only recently entered the $10^5$-compound regime \cite{enveda180, uritboonthai2026metlin}: broader benchmarking would therefore require non-open datasets or synthetic libraries (Appendix~\ref{app:coconut}).

\section{Conclusion}
\looseness=-1
We introduce an exact, scalable extension of myopic MCES that reduces the public-union problem to $1.15\times10^{7}$ integer programs, 0.04\% of all molecular pairs, while preserving a lossless connected-component partition.
Applying it at scale reveals that MCES-10 is not invariant to dataset size: as libraries grow, percolation collapses the achievable holdout and enriches the survivors in high molecular mass.
Splits built for \textit{de novo} elucidation benchmarks now conflate structural novelty with extrapolation across molecule size and fragmentation regime.
This may not be specific to MCES,
as percolation follows from single-linkage thresholding of any distance,
and the mass bias from thresholding on an \textit{absolute} one.
A fixed threshold therefore does not define a fixed benchmark,
but one that drifts with the size of the library it is applied to.
Crucially, a main premise of molecular machine learning is that
such an increase in library size and broader coverage of chemical space improves model generalization.
However, the measurement of this generalization improvement is increasingly impacted as the holdout size decreases and its mass bias increases.
The existing benchmark procedure therefore does not provide an invariant measure of this generalization improvement as today's chemical libraries grow in size.
Instead, structure-disjoint benchmarks must adopt criteria whose meaning is stable under scale. 
\section{Acknowledgments \& Disclosure of Funding}
We thank all members of Matterworks, Inc. for their advice and help on completing this work.
All authors of this work are employees or shareholders of Matterworks, Inc.

\bibliographystyle{unsrtnat}
\bibliography{reference}

\newpage
\appendix
\section*{Use of large language models}
Claude Opus 5 was used to aid in writing and implementation of code, including the vectorized bound cascade, the dispatch harness, and the plotting scripts that render figures from computed results. It was also used for copy-editing and prose polishing of the manuscript.
 
\section{The distance and the bounds}

\subsection{The distance}
 \label{app:distance}
A molecule is its heavy-atom graph: vertices are heavy atoms carrying an element label, edges are bonds carrying a weight $w(e)$ (single $1$, aromatic $1.5$, double $2$, triple $3$). $W(G)$ represents the total bond weight. \citet{kretschmer2025coverage} define:
\begin{equation}
\label{eq:mces}
d(A,B) = W(A) + W(B) - 2\,W\bigl(\mathrm{MCES}(A,B)\bigr)
\end{equation}
where the maximum-weight common edge subgraph is taken over correspondences between the two graphs induced by an injective, element-preserving partial map on atoms, and a matched bond pair contributes $\min(w_A, w_B)$. They also give an equivalent edit-distance form, which is the one used below. Such a map pairs up
some bonds and leaves the rest unmatched, and
\begin{equation}
    d(A,B) \;=\; \min \Big[
    \underbrace{\textstyle\sum_{\text{unpaired}} w}_{\text{bonds present in one molecule only}}
    \;+\;
    \underbrace{\textstyle\sum_{\text{paired}} |w_A - w_B|}_{\text{bonds present in both, differing in order}}
  \Big]
  \label{eq:distance}
\end{equation}
minimized over all such maps. The two agree termwise via $a + b - 2\min(a,b) = |a-b|$.
Equation~\ref{eq:distance} is what the bounds lower-bound and what the integer program minimizes.

Two conventions matter for what follows. Bond weights are read after the standardization pipeline (Appendix~\ref{app:standardization}), with aromaticity perceived identically for both inputs; otherwise a kekulized and an aromatic encoding of the same ring would give $d(A,A) \neq 0$. And every weight is a multiple of $\tfrac{1}{2}$, so all three bounds are exactly representable in binary floating point and the strict comparisons below carry no tolerance.

\subsection{The three vectorized bounds}
\label{app:bounds}

A bound is \textit{admissible} if it never exceeds $d$; a pair is discarded only when a bound proves $d \ge T$, so an overstatement would drop a genuinely near pair and leak. Each bound below is admissible because it lower-bounds the edit cost of \textit{every} correspondence, maximal or not, and hence of the minimizing one.

\paragraph{1. Total bond weight, $|W(A) - W(B)|$.} Splitting both totals by the correspondence and applying the triangle inequality (weights are non-negative, so the unmapped sums need no absolute value) bounds $W(A) - W(B)$ by the edit cost. Evaluated as a range query: with $W$ sorted once, all partners of $i$ within $T$ form a contiguous slice located by two binary searches, so this stage never materializes a per-pair value.

\paragraph{2. Per-class bond-weight imbalance, $\sum_\ell |a_\ell - b_\ell|$.} Let the class of a bond be the \textit{unordered} pair of its endpoint element labels, and $a_\ell$, $b_\ell$ the total weight of each molecule's bonds in class $\ell$. Element labels are preserved by the correspondence, so a bond can only be matched within its own class; unorderedness is needed because a bond may be matched with its endpoints exchanged. The edit cost therefore splits into independent per-class problems, and bound 1 applied within each class gives $|a_\ell - b_\ell|$. Summing proves admissibility. The same argument shows $|W(A) - W(B)| \le \sum_\ell |a_\ell - b_\ell|$, so bound 2 is strictly stronger. Bound 1 is just faster, being a range query.

The class label must contain nothing the correspondence fails to preserve, and refining it can only raise the computed value, so an inadmissible label fails silently, in the leaking direction. Bond order is the tempting case, since a matched pair is explicitly allowed to cost $|w_A - w_B| > 0$. Table~\ref{tab:class-label} shows the failure on benzene and cyclohexane. Atom degree is also inadmissible for the same reason, as an atom's degree in the common subgraph is generally lower than its parent. 

\begin{table}[h]
    \centering
    \caption{\textbf{Bond order must not enter the class label.} Benzene and cyclohexane have the same six-carbon skeleton and differ only in bond order, so their true distance is $3.0$: match all six bonds, each costing $|1.5 - 1.0| = 0.5$. Under the correct label both molecules occupy the single class \texttt{C-C}, their totals meet, and the bound is exactly $d$. Folding bond weight into the label splits them into disjoint classes in which each molecule holds zero weight in the other's, so the $l_1$ sum accumulates where it should cancel, overstating $d$ five-fold. Any threshold in $(3, 15]$ would then discard a pair at distance $3$.}
    \label{tab:class-label}
    \begin{tabular}{lccc}
        \toprule
         & \multicolumn{1}{c}{endpoint elements} & \multicolumn{2}{c}{elements $+$ bond weight} \\
        \cmidrule(lr){2-2} \cmidrule(lr){3-4}
        class & \texttt{C-C} & \texttt{C-C:1.5} & \texttt{C-C:1.0} \\
        \midrule
        benzene $a_\ell$      & $9.0$ & $9.0$ & $0.0$ \\
        cyclohexane $b_\ell$  & $6.0$ & $0.0$ & $6.0$ \\
        \midrule
        $|a_\ell - b_\ell|$   & $3.0$ & $9.0$ & $6.0$ \\
        bound $\sum_\ell |a_\ell - b_\ell|$ & $\mathbf{3.0}$ & \multicolumn{2}{c}{$\mathbf{15.0}$} \\
        \bottomrule
    \end{tabular}
    
\end{table}

\paragraph{3. Per-element weighted-degree sequences, \texttt{filter1}.} For each element, take the weighted degrees ($\deg(u) = \sum_{e \ni u} w(e)$) of that element's atoms in each molecule, sort both descending, zero-pad to a common length, and sum the $l_1$ distances across elements; halve the total. Two facts make this admissible. A bond edit perturbs the weighted degree of at most two atoms, so the total degree discrepancy between corresponding atoms is at most twice the edit cost, and the factor of $\tfrac{1}{2}$ is exactly the handshake identity $\sum_u\deg(u) = 2W(G)$, correcting for each bond being counted at both endpoints. Sorting both sequences in the same direction is the optimal one-dimensional coupling, so the computed value is a \textit{minimum} over pairings rather than the cost of one arbitrary alignment; the zero-padding makes a partial matching, in which an unmatched atom pays its full degree, equivalent to a perfect one. For the vectorized re-implementation we are careful that (1) both sequences are sorted in the \textit{same} direction (opposite directions maximize $l_1$ instead) and (2) elements are kept separate (pooling is still admissible, since a larger feasible set lowers the minimum, but is strictly weaker).

\paragraph{The bounds are incomparable.} Neither of bounds 2 and 3 dominates the other, so both run. For benzene vs. cyclohexane all three give $d = 3.0$; for \texttt{CCO} vs benzene, bound 2 gives $d=9.0$ while \texttt{filter1} gives $8.0$. Pairs where \texttt{filter1} wins also occur. 

The entire bound cascade is checked exhaustively over the MassSpecGym distance matrix to ensure validity (see Appendix~\ref{app:validation}).

\newpage
\section{The pipeline}
\label{app:implementation-details}

The numbers in this section are for the full public union at $T=10$ unless stated otherwise. The bounds themselves are Appendix~\ref{app:bounds}. 

\subsection{Candidate generation} 
The three vectorized bounds are evaluated over all $\binom{N}{2}$ pairs without ever materializing them. Bond weights $W$ are sorted once; for each structure $i$ the partners satisfying $|W(A) - W(B)| < T$ form a contiguous slice of that order, located by two binary searches, so bound 1 is a range query rather than a comparison. Only the pairs inside those bands are enumerated, and bounds 2 and 3 are applied to them as batched array operations, bound 3 against a zero-padded per-element degree matrix. On the union the three stages reduce $25{,}921{,}254{,}205$ pairs to $18{,}525{,}485{,}344$, then $3{,}634{,}287{,}497$, then $3{,}351{,}843{,}806$ candidates ($12.93\%$).

Each surviving pair is assigned to one of $20$ buckets by its \texttt{filter1} value, in steps of $0.5$ from $0$ to $9.5$. There is no bucket at 10 as \texttt{filter1} is a lower bound and the near test is strict, so a value of exactly $T$ already proves $d \ge T$ and the pair is discarded. Candidates are written to disk per bucket in shards, so the working set is one shard rather than the whole candidate array.

\subsection{Dispatch} 
Buckets are walked in ascending order, with the cheapest bound first, and the bucket is consumed in blocks. Before a block is dispatched, every pair in it is tested against a union-find over \textit{proven-near} edges, and pairs whose endpoints already share a component are skipped. The block is then solved in parallel, and its merges applied afterwards, so every skip test within one block sees the same forest. That snapshot property is what allows the test to be a single batched root chase rather than a per-candidate find; moving a merge inside the skip loop would make the batched form skip fewer pairs than the scalar form, i.e. change the solve set. On the union the skip removes $3{,}277{,}225{,}708$ of the $3{,}351{,}843{,}806$ candidates ($97.77\%$), leaving $74{,}618{,}098$ dispatched pairs (Table~\ref{tab:funnel}).

\begin{figure}
    \centering
    \includegraphics[width=\linewidth]{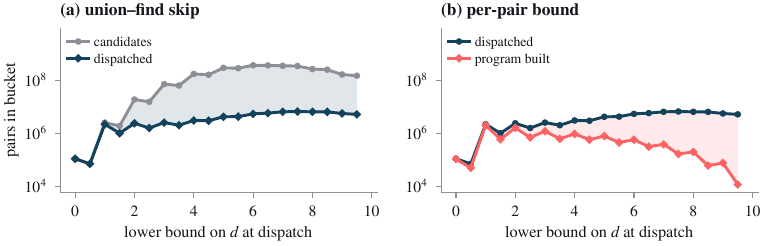}
    \caption{\textbf{Both per-pair reductions strengthen along dispatch order.} Candidates are dispatched cheapest-first, in ascending order of their vectorized bound. Together they concentrate integer programs to the cheaper end of the dispatch order, which is also the end that builds components. The last 4 buckets hold 31.8\% of dispatched pairs but only 3.0\% of the programs. (a) The union-find skip cannot skip at all in the first few buckets as no components exist yet, but removes 96\% of the last bucket, because by then the giant has absorbed most of the library and a candidate's endpoints are usually already connected. (b) The per-pair bound's rejection rate climbs because a pair whose weaker bound already sits near $T$ is exactly the pair a stronger bound most easily pushes past. It rises from skipping 2.5\% of pairs to 99.8\% of them}
    \label{fig:bucket}
\end{figure}

Ascending-bound order is a throughput heuristic, but not a correctness requirement. It forms components early and so makes more later pairs skippable. Its effect is visible per bucket: the first two buckets skip nothing at all, because no components exist yet, and the last skips $96.5\%$ (Figure ~\ref{fig:bucket}).

Block size trades skip quality against parallelism, since pairs inside one block cannot benefit from each other's results. Measured on this codebase, the end-to-end saving is $6.45\times$ at block size 1, $6.27\times$ at 50,000 and $5.01\times$ at 1,000,000; the runs reported here used 2,000,000, chosen for worker occupancy at 240 processes.

\begin{table}[h]
\centering
\small
\caption{The pipeline on the public spectral union, $N = 227{,}690$, $T = 10$. What each test removes. Stage 1 and 2 are vectorized bounds that we introduce; Stage 3 is a vectorized version of \texttt{filter1}; Stage 4 is our union-find based skipping we introduce; Stage 5 is the per-pair neighborhood bound (\texttt{filter2}), applied immediately before the solver. 99.956\% of pairs are pruned before ever touching a solver.}
\label{tab:funnel}

\begin{tabular}{@{}clrrr@{}}
    \toprule
    & \textbf{Test} & \textbf{Pairs removed} & \textbf{Pairs left} & \textbf{Saving} \\
    
    \midrule
      & all pairs $\binom{N}{2}$       & ---                  & 25{,}921{,}254{,}205 & --- \\
    \midrule
    
    \multicolumn{5}{@{}l}{\emph{vectorized, over all pairs}} \\
    1 & total bond weight              &  7{,}395{,}768{,}861 & 18{,}525{,}485{,}344 & $1.40\times$ \\
    2 & per-class weight imbalance     & 14{,}891{,}197{,}847 &  3{,}634{,}287{,}497 & $5.10\times$ \\
    3 & per-element degree sequences   &      282{,}443{,}691 &  3{,}351{,}843{,}806 & $1.08\times$ \\
    
    \multicolumn{5}{@{}l}{\emph{per pair}} \\
    4 & endpoints already co-clustered &  3{,}277{,}225{,}708 &      74{,}618{,}098  & $\mathbf{44.92\times}$ \\
    5 & neighbourhood bound, per pair  &       63{,}141{,}783 &      11{,}476{,}315  & $6.50\times$ \\
    \midrule
      & \textbf{end to end}            & \textbf{25{,}909{,}777{,}890} & \textbf{11{,}476{,}315} & $\mathbf{2{,}259\times}$ \\
    \bottomrule
\end{tabular}

\end{table}

\subsection{The per-pair solver} 
A dispatched pair is first tested against \texttt{filter2}, the neighborhood bound of \citet{kretschmer2025coverage} which is computed through a bipartite matching and does not vectorize over $10^{10}$ pairs; it is therefore applied here, per pair, and nowhere earlier. It proves $63{,}141{,}783$ of the $74{,}618{,}098$ dispatched pairs far ($84.62\%$) without an integer program, leaving $11{,}476{,}315$ programs actually constructed.

We use myopic-MCES's integer program unmodified \cite{kretschmer2025coverage}. The property we rely on is that when a threshold is set the program carries it as the hard constraint $\sum_e w_e c_e \le T$ on the objective, so its feasible region is exactly the set of correspondences of cost at most $T$: an integer-feasible solution witnesses $d \le T$ even when optimality is unproven, and a proof of infeasibility witnesses $d > T$. It is given a 5s wall-clock budget and solved in process by HiGHS with \texttt{threads=1} and \textbf{presolve disabled}. Presolve is off for correctness, as on threshold-constrained instances stock HiGHS reports \texttt{infeasible} for pairs that are demonstrably near. Since infeasible is read here as a proof of far-ness, we disable it as it would be a leak.

Three exits are possible for the solver: \textbf{(1) Optimal.} The program proves an optimum, giving $d$ exactly. \textbf{(2) Infeasible.} The program proves no correspondence achieves cost $\le T$, hence $d > T$. This is a proof of \emph{far}, not an undecided outcome, and most programs end this way (Table~\ref{tab:ledger}). \textbf{(3) Timeout.} A budget expiry is reported as a single outcome by the solver, but it covers two cases that change what is resolved. If the search holds a solution whose optimality is unproven, that solution is a valid correspondence of cost at most $T$, so the pair is certified near. If it ends having no solution at all, no proof of verdict is exhibited and the pair is merged near, in the leakage safe direction.


\begin{table}[h!]
    \centering
    \small
    \caption{\textbf{The exit outcomes of the 11,476,315 pairs that survived every pruning measure.} Infeasibility proves farness, by construction of the program. The near verdicts form a graph of 744,758 edges, of which 220,598 merge components and 524,160 close cycles, for a total of 7,092 clusters with a giant of 217,440 (95.50\%). The dispatch stage took 3,813 core hours, against 58,901 for myopic-MCES under MassSpecGym's settings. }

    \begin{tabular}{@{}lrrl@{}}
        \toprule
        \textbf{How the integer program terminates} & \textbf{Programs} & \textbf{Share} & \textbf{Verdict} \\
        \midrule
        proves infeasibility                & 10{,}531{,}154 & $91.76\%$ & far \\
        proves an optimum, $d \ge T$        &      200{,}403 &  $1.75\%$ & far \\
        proves an optimum, $d < T$          &      689{,}280 &  $6.01\%$ & near \\
        budget expires, unproven solution   &        2{,}415 &  $0.02\%$ & near \\
        budget expires, no solution found   &       53{,}063 &  $0.46\%$ & near, conservatively \\
        \midrule
        \textbf{total}                      & \textbf{11{,}476{,}315} & $100\%$ & \\
        \bottomrule
    \end{tabular}
    \label{tab:ledger}
\end{table}

\subsection{The edge rule and re-solving timeouts} 
\label{app:re-solve_timeouts}
A pair is near if and only if it timed out, or was solved exactly with $d < T$. Undecided pairs are banked near because the two errors are not symmetric: wrongly separating a near pair leaks, whereas wrongly merging a far pair only costs holdout.

Because a timed-out pair is banked near, a reported giant fraction is an \textit{upper} bound. Timeouts can be recovered exactly from the banked rows and re-solved at a longer budget. These timed-out edges mostly do not change the partition: by contracting the proven-near edges first, a timeout whose endpoints already share a proven component cannot change the partition however it resolves. We find that on the union it leaves 3,265 of 55,478 timeouts (5.89\%) to be re-solved. Solving these to completion moves the giant from 95.4983\% to 95.3441\% (0.154 percentage points), and the mass AUC from 0.8810 to 0.8853 (+0.0043). 

\subsection{Every rung and its lower bound}
\label{app:bracket}
A timed-out pair is banked near so every giant fraction is an upper bound, and the timeout count does grow with library size (Table~\ref{tab:bracket}). What governs the interval is not how many pairs timeout but how many of them bridge two \textit{proven} components; the rest cannot change the partition however they resolve. Both ends come from the same banked verdicts with no further solving. The upper end treats every timeout as near, which is what we report; the lower end treats every timeout as far, which is the leaky assignment; these two bound the possible size of the giant component. 

The interval never exceeds $0.58$ percentage points, so the trend is not an artifact of a budget that bites harder at scale. It also survives the pessimistic reading: taking the lower end at every rung, the giant still climbs monotonically from $63.02\%$ to $95.33\%$.

\begin{table}[h]
\centering
\caption{\textbf{Every rung of the ladder is bounded.} Timeout counts grow with $N$ while the interval does not. The public union at $T=10$; rungs with several seeds are averaged and the interval is the widest of them.}
\label{tab:bracket}
\begin{tabular}{rrrrr}
\toprule
$N$ & timeouts & giant, all far & giant, all near & interval \\
\midrule
$10{,}000$  &      $993$ & $63.02\%$ & $63.45\%$ & $0.58$ pp \\
$20{,}000$  &   $2{,}587$ & $76.89\%$ & $77.13\%$ & $0.30$ pp \\
$40{,}000$  &   $5{,}652$ & $86.08\%$ & $86.22\%$ & $0.17$ pp \\
$80{,}000$  &  $12{,}764$ & $91.31\%$ & $91.38\%$ & $0.08$ pp \\
$160{,}000$ &  $32{,}198$ & $94.41\%$ & $94.51\%$ & $0.17$ pp \\
$227{,}690$ &  $55{,}478$ & $95.33\%$ & $95.50\%$ & $0.17$ pp \\
\bottomrule
\end{tabular}
\end{table}

\subsection{The pipeline computes the dense partition}
\label{app:union_skip}

Clusters are the connected components of the near graph, so it suffices that neither pruning step removes a near edge that could change those components. Every bound is a lower bound on $d$, so a bound reaching $T$ proves $d \ge T$ and the pair is far. All three tests are strict and no near pair is ever discarded. The union-find skip removes a pair only when its endpoints already lie in one component of the \textit{confirmed-near} graph, which tracks edges that are \textit{proven} near. An edge created by a solver timeout is not confirmed, so no pair is ever skipped on the unresolved verdict. This is also what makes the timeout re-solve of Appendix~\ref{app:re-solve_timeouts} sound, as the timeouts that could still matter are exactly those bridging two proven components.


The pruning is lossless, and keeps the same leakage guarantee. The partition is therefore exactly the one a dense computation over all $\binom{N}{2}$ pairs would produce, \textit{given the same solver configuration}. This gives an upper bound on the giant component. Appendix~\ref{app:bracket} bounds it at every scale, and Appendix~\ref{app:validation} settles it outright on MassSpecGym: re-solving until nothing is left unproven recovers their published partition exactly. 

\subsection{Redundant solves, and speedup reliance on percolation}
\label{app:redundancy}

The near graph is not a forest. On the union, 744,758 near verdicts produce only 220,598 merges, and so 70.4\% of near edges close cycles rather than connect anything. The skip is tested when a pair is \textit{dispatched}, not when it returns, and a block is solved in parallel, so two pairs in one block can come back near across the same cut while only the first merges. The merging edges are therefore not a \textit{minimum} spanning forest: dispatch orders by \texttt{filter1}
 bucket rather than by true distance, and a bucket spans a range of distances, so the forest actually built has total weight $1,345,783.5$ against $1,124,467.0$, 19.7\% heavier. Neither fact affects the partition, which does not depend on which spanning forest is found.

 Union-find admits at most $N-1$ merges on any corpus, and here 220,598 of a possible 227,689 are used (96.9\%). That near-saturation is equivalently percolation, and it is also precisely what makes union-find skipping fast. The skip's saving grows because the giant grows, and so the method is fastest exactly on the libraries where the split is least usable. The redundancy is pure waste (524,160 solves that change nothing about the partition), and a smaller dispatch block would recover part of it at the cost of parallelism; we did not tune it. 

\newpage
\section{Validation against MassSpecGym}
\label{app:validation}

MassSpecGym publishes an all-pairs myopic-MCES matrix over its 34,731 structures (603,103,815 pairs), used here as an external oracle. Its entries carry the same three-way semantics as ours: below the threshold is an exact distance, exactly 10.0 is the integer program returning infeasible, larger is whatever bound proved the pair. 

\paragraph{Bounds and dispatcher.} Substituting the matrix for the solver isolates everything but the integer program. Exhaustively over the full set, every one of the 1,710,273 below-threshold entries survives the cascade, with zero dropped. The union-find partition is identical to resolving every candidate.

\paragraph{Partition.} At the 5s budget our partition is coarser, as the conservative rule requires: 7,659 clusters and a giant of 23,427 against their 7,770 and 22,768. Resolving the 13,437 bridging timeouts without a budget yields 731 near and 12,706 far, with none left unproven. The partition is then \textit{identical} to MassSpecGym's at $7{,}770$ and a giant of $22{,}768$. Across the $125{,}803$ pairs we prove near, none is recorded at or above the threshold by them, and across the $25{,}272$ we prove far, none is recorded below it, and their distances are bit-identical.

\paragraph{Bound equivalence.} Admissibility is a property of a bound and is proved in Appendix~\ref{app:bounds}; agreeing with the published \textit{implementation} is a property of the code. Our vectorized \texttt{filter1} was checked differentially against \texttt{myopic-mces}'s on 500,000 pairs drawn uniformly from the union, from our packed descriptors on one side and the \texttt{networkx} graphs on the other. The two agree exactly on every pair, maximum absolute difference of 0, over bound values from 0 to 89. 

\paragraph{The myopic-MCES baseline.} \label{app:baseline}
Myopic-MCES has no candidate generation and no dispatch skip, so splitting a library with it means calling its distance function on every pair. Its cost was measured, by timing the published function on $300{,}000$ pairs drawn uniformly from the union with the solver configured as ours: $13.844 \pm 0.108$ ms per call, or $100{,}774$ core-hours over the union's $2.59\times10^{10}$ pairs. MassSpecGym builds its splits with \texttt{always\_stronger\_bound=False}, which computes \texttt{filter1} first and short-circuits when it already exceeds $T$; that is $8.180$\,ms per call and $58{,}901$ core-hours. Dispatch counts were measured per rung over $5\times10^{6}$ sampled pairs, anchored by an exhaustive enumeration of all $49{,}995{,}000$ pairs at $N = 10{,}000$ which the sample reproduces to $0.091\%$: a near-constant $3.00$--$3.11\%$ of pairs across a $23\times$ range in $N$, hence $N^{2.01}$ against $N^{0.93}$ for ours, and $803{,}175{,}246$ programs against $11{,}476{,}315$ at the full union.

\newpage
\section{Experimental details}
\label{app:exp_setup}

\subsection{Compound sets.} 
Structures were drawn from two public spectral-library deposits, SpectraVerse 1.0.1 \cite{spectraverse2025} (Zenodo \texttt{10.5281/zenodo.17252773}) and Enveda-180 \cite{enveda180}  (Zenodo \texttt{10.5281/zenodo.21346580}).

\textbf{Exactly one field is read from each deposit.} From SpectraVerse we take the \texttt{SMILES} field; from Enveda-180 we take \texttt{smiles}, discarding null rows. Only unique entries are taken, so a structure measured under many adducts, collision energies, or instruments enters once (molecular splitting is invariant to these). Nothing else is consumed: no peak lists, precursor masses, adducts, collision energies, instrument or ionization metadata, retention or ion-mobility values, compound names, or database identifiers. This study needs only the structure set.

Two further quantities are \textit{derived} from that column and used downstream. The canonical SMILES produced by standardization is the identity key on which the union is deduplicated. The heavy-atom graph parsed from it is the sole input for the distance (Appendix~\ref{app:distance}). Monoisotopic exact mass, reported throughout the paper, is computed from the standardized structure from RDKit.

\subsection{Standardization}
\label{app:standardization}
The criterion reads the heavy-atom graph, so any difference in how a compound is \textit{recorded} is a difference in its distance. This matters because the two deposits were curated independently and record the same compound under different registration conventions. A pair of records that standardization would collapse but does not is a harmless duplicate: two encodings of one compound at $d=0$ merge and cost little holdout, but an encoding difference large enough to reach $T$ certifies two records of the same compound \textit{far}, and they are then free to land on opposite sides of the split. That is precisely the leakage the split exists to prevent, and it is not a small effect. Salt forms are the clear case, since a disconnected counter-ion contributes its entire bond weight as unpaired: chloride carries none, but maleate carries $10.0$, tartrate $11.0$, and citrate and besylate $14.0$, each at or beyond $T=10$ on its own. A free base and its tartrate salt would therefore be proven far by bound 1 alone, without a solver ever seeing them.

Stereochemistry is invisible to an element-labeled graph, so stereoisomeric records already sit at $d=0$; stripping it removes duplicate nodes and computations rather than changing any verdict. Tautomer and aromaticity conventions must be applied to both inputs for the reason given in Appendix~\ref{app:distance}: a kekulized and an aromatic encoding of one ring would otherwise give $d(A,A) \neq 0$. 

We re-purpose SpectraVerse's own \texttt{step2-3\_standardization.py} pipeline so that compound identity in this study matches theirs. In order: SMILES are stripped and then the base SMILES is parsed; stereochemistry removed; sanitized and explicit hydrogens dropped; \texttt{rdMolStandardize.Cleanup} disconnects metals and performs reionization, followed by \texttt{FragmentParent}, to desalt; charge-separated sulfur and phosphorus bonds are repaired; charges are neutralized by \texttt{Uncharger}, then nine SMARTS patterns covering imidazoles, amines, carboxylates and alcohols, thiols, sulfonamides, enamines, tetrazoles, sulfoxides, and amides, then by \texttt{Uncharger} again; sanitized; tautomers canonicalized with \texttt{TautomerEnumerator}; and finally round-tripped through canonical non-isomeric SMILES, since \texttt{Canonicalize} can leave an implicit-hydrogen state that differs from a freshly parsed molecule.

\subsection{The rung grid}
\label{app:rung}
Each arm is measured on a ladder of library sizes. A rung at size $n < N$ is a uniform sample without replacement of $n$ structures from that arm's pool, drawn with \texttt{numpy.random.default\_rng(1000 + s)} for seed index $s$ and then sorted, so a rung is reproducible from its arm and seed alone. Rungs at the full library size involve no sampling. 

\begin{table}[h]
\centering
\caption{\textbf{Every run reported.} Seeds are $1000+s$ for $s$ in the listed range; a full-library rung is unsampled and unseeded. All runs use $T=10$, a $5$\,s per-program budget, 240 workers and a dispatch block of $2{,}000{,}000$, except the threshold sweep as noted.}
\label{tab:rungs}
\small
\begin{tabular}{llll}
\toprule
Arm & Pool & Rungs $n$ & Seeds $s$ \\
\midrule
Public union            & 227,690 & 10{,}000 / 20{,}000 / 40{,}000 & $0,1,2$ \\
                        &         & 80{,}000 / 160{,}000           & $0,1$ \\
                        &         & 227{,}690 (full)               & --- \\
Enveda-180 alone        & 183,476 & 10{,}000 / 20{,}000 / 40{,}000 & $0,1,2$ \\
                        &         & 80{,}000 / 160{,}000           & $0,1$ \\
                        &         & 183{,}476 (full)               & --- \\
\bottomrule
\end{tabular}
\end{table}

\paragraph{Each rung is solved from scratch.} No rung is obtained by restricting a larger run's output, and none can be. The banked verdicts are a \textit{spanning} structure: the dispatch skip never solves a pair whose endpoints are already connected, so a smaller set of structures may need edges that were never proven in the near graph of the full library. Restricting would therefore under-count near edges, and always in the leaking direction.

\subsection{Run configuration}

Every split in this paper uses the same solver settings: threshold $T=10$ (the value MassSpecGym adopts, varied only in Appendix~\ref{app:threshold}), a $5$\,s wall-clock budget per integer program, HiGHS in process with \texttt{threads=1} and presolve disabled for the correctness reason in Appendix~\ref{app:implementation-details}, and timed-out pairs banked near. Candidates are bucketed into $20$ bins of width $0.5$ on \texttt{filter1} and dispatched cheapest-first in blocks of $2{,}000{,}000$ across 240 worker processes. The full-union run at these settings takes $57{,}193$\,s of wall clock ($15.89$\,h).

Two quantities are reported throughout. \textit{Achievable holdout} is $1 - g/N$ for a giant component of $g$ structures: the giant is indivisible, so this is the largest evaluation set the cluster structure could ever permit, whatever assignment procedure is applied afterwards. \textit{Mass holdout AUC} is the Mann--Whitney statistic for monoisotopic exact mass separating the giant from its complement; $0.5$ is no mass difference, and higher means the compounds outside the giant are heavier.

\subsection{The COCONUT arm}

The third library of Appendix~\ref{app:coconut} draws on the COCONUT 09-2024 dump \cite{coconut}, Zenodo record \texttt{13692394}, standardized by the identical pipeline. One filter is applied to it: structures at or above $1000$\,Da are dropped, since the study concerns small-molecule spectrum-to-structure prediction and instruments in these deposits rarely operate above that mass. No structural filter is applied to any source.

SpectraVerse's 44,240 structures are held fixed and the 183,450 structures Enveda-180 contributes exclusively to the union are replaced by the same number drawn uniformly from the filtered COCONUT pool, with seed \texttt{20260820}. The result is a library of the same size as the public union, differing from it in exactly one respect. Its rungs are random subsamples on the same grid as the main ladder rather than nested prefixes, so a comparison at matched $n$ is like-for-like.

\subsection{Hardware and software} 
One machine, 384 cores, 1,511 GB RAM; splits computed using 240 worker processes and the measurement sweeps of Appendix~\ref{app:validation} used 220. Python 3.10.12, NumPy 2.2.6, RDKit 2026.03.4, PuLP 3.3.2, \texttt{highspy} 1.15.1, \texttt{myopic-mces} 1.2.2.

\newpage
\section{The size filter is the criterion}
\label{app:band}

Bound 1 (i.e. total bond weight) additionally gives us a characterization for which molecules can be linked at all. Since $|W(A) - W(B)|$ lower-bounds $d$, any pair whose total bond weights differ by $T$ or more is far, so a molecule's possible partners lie in a window of \textbf{fixed absolute width $2T$} around its own bond weight. For a molecule with $W=14$ that window spans $\pm 71\%$ of its own weight, while at $W=83$ it spans $\pm 12\%$. This allows us to formulate an exact factorization of the chance that a uniformly drawn library member is near a given molecule:

\begin{equation}
\label{eq:band-factor}
P(\text{near}) \;=\; \underbrace{P(\text{in band})}_{\text{composition}} \;\times\;
                     \underbrace{P(\text{near} \mid \text{in band})}_{\text{criterion}},
\end{equation}

where the two factors answer different questions. The first asks how many molecules happen to sit at a similar weight, which is a property of the library composition. The second asks how many of those that are similar in weight, are actually near, which is the criterion itself: clearing $ d < T$ means sharing all but $T$ of the combined weight, so the relative tolerance $T/2W$ tightens as molecules grow.

\begin{figure}[t]
\centering
\includegraphics[width=.58\linewidth]{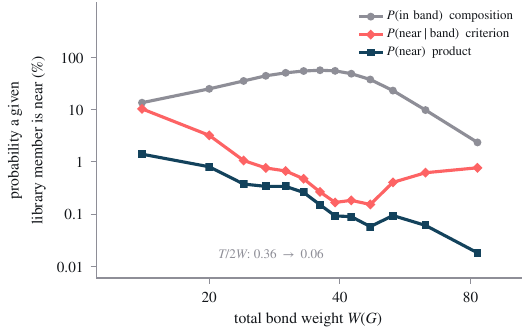}
\caption{Probability that a uniformly drawn library member is near a molecule of the given total bond weight on MassSpecGym's complete matrix, split by Equation~\ref{eq:band-factor}. Because of the fixed absolute width, the probability a library member has a near neighbor falls with total bond weight.}
\label{fig:band}
\end{figure}

We evaluate these factors on MassSpecGym's published all-pairs matrix rather than our own verdicts as that matrix is \textit{complete}, with every one of the 603,103,815 pairs carrying a distance, which makes the near-degree of every structure exact. 

Over the range holding most of the library, \textit{composition} opposes the effect, but the criterion overwhelms it. This is the sense in which the bias follows from thresholding an absolute distance and no choice of library escapes it.

\subsection{The collapse is a function of the chosen threshold}
\label{app:threshold}

Everything reported at $T=10$ follows MassSpecGym's choice. Since the mechanism above is that an \textit{absolute} threshold becomes a size filter, the threshold is the one parameter that should move the result, and it does. Re-running the full public union at $T=5$ with every other setting fixed:

\begin{center}
\begin{tabular}{lrr}
\toprule
 & $T = 10$ & $T = 5$ \\
\midrule
clusters              & $7{,}092$   & $85{,}602$ \\
giant                 & $95.50\%$   & $26.54\%$ \\
achievable holdout    & $4.50\%$    & $\mathbf{73.46\%}$ \\
mass holdout AUC      & $0.881$     & $\mathbf{0.637}$ \\
\bottomrule
\end{tabular}
\end{center}

\paragraph{A representative split.} At $T=10$ an 80/10/10 split does not exist. The giant holds $95.5\%$ of the library and is indivisible, so whichever fold receives it overshoots to $95.5\%$ whatever ratios are requested, and no assignment procedure can do better. At $T=5$ the giant is $26.5\%$, the ratios become achievable, and folds can actually be shown. Routing whole clusters largest-first to the fold furthest below its target gives $182{,}152$ / $22{,}769$ / $22{,}769$ structures, which is $0.800$ / $0.100$ / $0.100$ to three decimal places.

\begin{figure}[t]
\centering
\includegraphics[width=\linewidth]{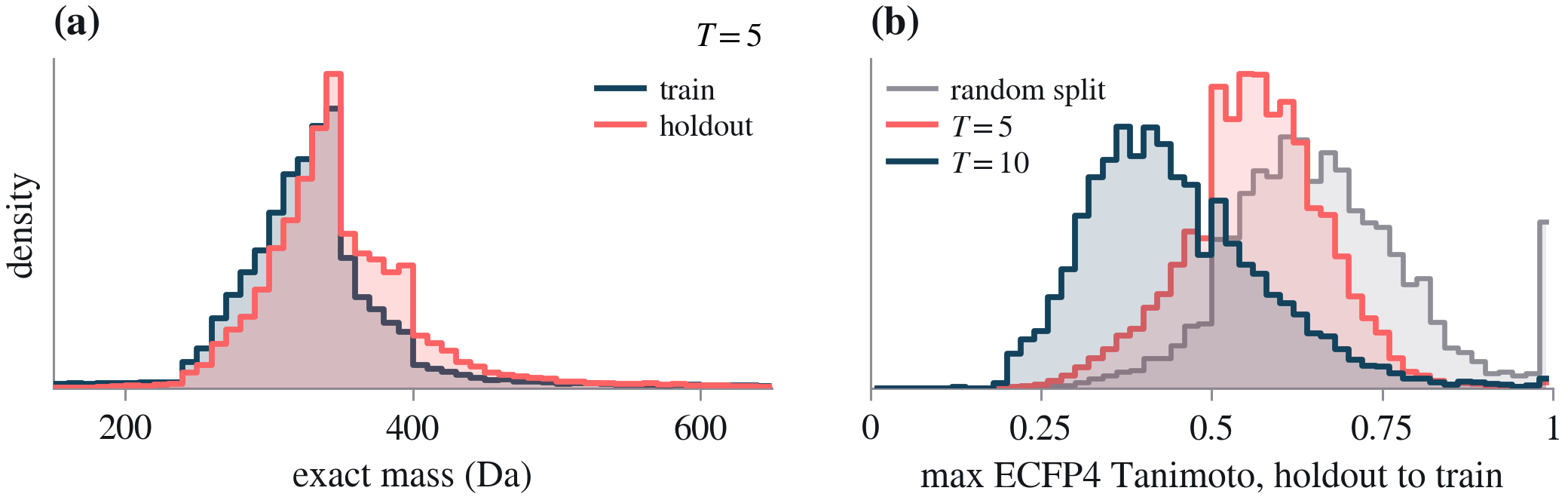}
\caption{\textbf{At a lower threshold a conventional split exists and is nearly mass unbiased, but its holdout is less novel.} (a) Train against holdout for an achievable 80/10/10 split of the public union at $T=5$. Median mass is 331.2 Da against 343.1 Da, compared to a $T=10$, where the giant component and the available holdout differ by 122.1 Da. (b) Maximum ECFP4 Tanimoto to any training compound. Medians are 0.427 at $T=10$, 0.566 at $T=5$, and 0.648 at random, so the stricter threshold does buy a novel holdout.}
\label{fig:t5}
\end{figure}

The mass bias is largely gone from that split (Figure~\ref{fig:t5}a). Median exact mass is $331.2$\,Da in train against $343.1$\,Da in the holdout, a shift of $11.9$\,Da, where the $T=10$ giant and its complement differ by $122.1$\,Da.

\paragraph{What a threshold does and does not buy.} The construction guarantees that no cross-fold pair is within MCES $T$. It does not guarantee the holdout is \textit{unfamiliar}, and those are different claims. Figure~\ref{fig:t5}b takes the maximum ECFP4 Tanimoto from each holdout compound to any training compound, for the largest evaluation set each threshold permits, against a random split of the same library at the same fold sizes.

Read together with the table, the panel prices the threshold on both axes at once. Lowering $T$ from $10$ to $5$ buys a factor of $16$ in achievable holdout and removes most of the mass bias, and costs novelty: the median max-Tanimoto rises from $0.427$ to $0.566$, and the fraction of holdout compounds with a training neighbor above $0.7$ from $4.1\%$ to $8.7\%$. Both remain far below the random control at $0.648$ and $34.9\%$, so both thresholds are doing real work; $T$ chooses where on that curve to sit.

Lowering the threshold is not a free fix. $T$ defines \textit{near}, and therefore defines the leakage the split exists to prevent, so a lower threshold buys usability by weakening the guarantee: pairs at MCES distance 5 to 10, which at these molecular sizes can be a single ring substitution, become free to straddle folds. This is visible through the max-Tanimoto of the holdout to train. What neither threshold removes is the tail at Tanimoto $1.0$. At $T=10$, $0.254\%$ of the holdout ($26$ compounds) is fingerprint-identical to a training compound; at $T=5$, $0.154\%$ ($70$ compounds).  

\subsection{Where fingerprint and MCES disagree most}
\label{app:fingerprint_disagree}

\begin{figure}[t]
    \centering
    \includegraphics[width=0.8\linewidth]{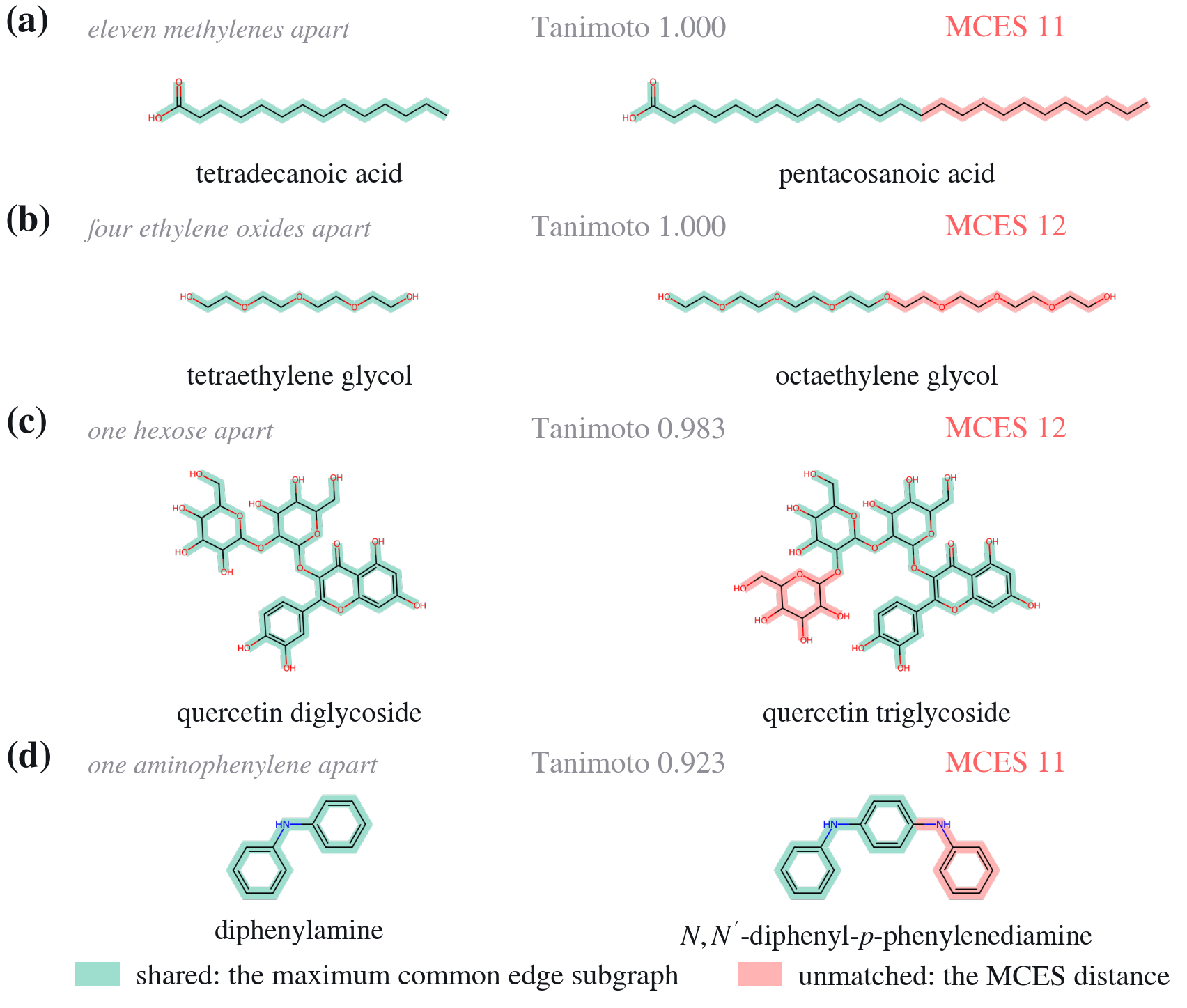}
    \caption{\textbf{Homologous pairs that are near-identical to ECFP4 yet provably distant
under MCES.} Green marks the maximum common edge subgraph; red marks the
unmatched bonds that constitute the MCES distance. (a)~Tetradecanoic vs.\
pentacosanoic acid and (b)~tetra- vs.\ octaethylene glycol are
fingerprint-identical (Tanimoto 1.000) despite MCES distances of 11 and 12;
(c)~quercetin di- vs.\ triglycoside (Tanimoto 0.983, MCES 12) and
(d)~diphenylamine vs.\ $N,N'$-diphenyl-$p$-phenylenediamine (Tanimoto 0.923,
MCES 11) show the same repeat-unit blindness at high but sub-unit Tanimoto.}
    \label{fig:pairs}
\end{figure}

On MassSpecGym's complete all-pairs matrix, 2,853 pairs attain ECFP4 Tanimoto
$\ge 0.999$ while being provably distant at MCES $\ge10$. These pairs span 953
structures across 129 disjoint families, a general property of the
representation rather than one pathological series. Their MCES distances run
from 11 to 109 (median 16), and $43.5\%$ of the pairs differ by an exact
multiple of CH$_2$. Tetradecanoic and pentacosanoic acid are the clearest case
(Figure~\ref{fig:pairs}a): eleven methylenes apart, yet fingerprint-identical,
with an MCES distance of 11, so a $T=10$ split is free to place them on
opposite sides.

The blind spot is one of counting. A fingerprint records a hashed set of
substructural environments, so molecules built from repeating units collapse onto the same bit vector.
Tetraethylene and octaethylene glycol are likewise fingerprint-identical,
despite sitting four ethylene oxide units and MCES 12 apart
(Figure~\ref{fig:pairs}b). A further 3,599 pairs fall between Tanimoto 0.9 and 1.0 while remaining provably distant.
The most fingerprint-similar of these, at 0.9833, is a pair of quercetin
glycosides differing by a single hexose, 162.05~Da of added sugar, MCES 12
(Figure~\ref{fig:pairs}c), and diphenylamine against its phenylenediamine
homologue reproduces the same failure in an aromatic system
(Figure~\ref{fig:pairs}d).

These are the exact failure modes that a split should prevent, and they leak through two distinct channels. 

The premise of molecular networking is that structurally related molecules fragment similarly, yet spectral similarity is only a loose proxy for structural similarity \cite{wang2016gnps, huber2021spec2vec}. The regime in which the proxy is most reliable is exactly the one these pairs occupy: a single additive modification on an otherwise intact scaffold, which leaves the core fragmentation ladder untouched and displaces the remaining peaks by a constant offset. A model trained on tetradecanoic acid has therefore already seen most of the spectrum that pentacosanoic acid produces. Splitting the two by MCES places a compound in the test fold whose fragmentation behavior the model was taught in training.

ECFP4 fingerprints are what the field has largely landed on as the intermediary prediction from spectra for retrieval or \textit{de novo}. MIST is a spectrum encoder trained to predict molecular fingerprints directly from spectra \cite{goldman2023mist}, and many recent state-of-the-art \textit{de novo} methods decode structures from a predicted fingerprint produced by an encoder of that lineage \cite{bohde2026frigid, neo2025molforge, bohde2025diffms}. For a fingerprint-identical pair split across folds, the encoder's supervision target for the held-out molecule is bit-for-bit the vector it was trained to emit for the training molecule. 

\newpage
\section{A third library: SpectraVerse with COCONUT}
\label{app:coconut}

Both arms in the main text draw on MS/MS reference deposits. It is fair to ask whether what we observe is peculiar to them, so we repeated the experiment on a third library assembled from a different source, and report it here as corroboration.

\begin{figure}[t]
\centering
\includegraphics[width=0.7\linewidth]{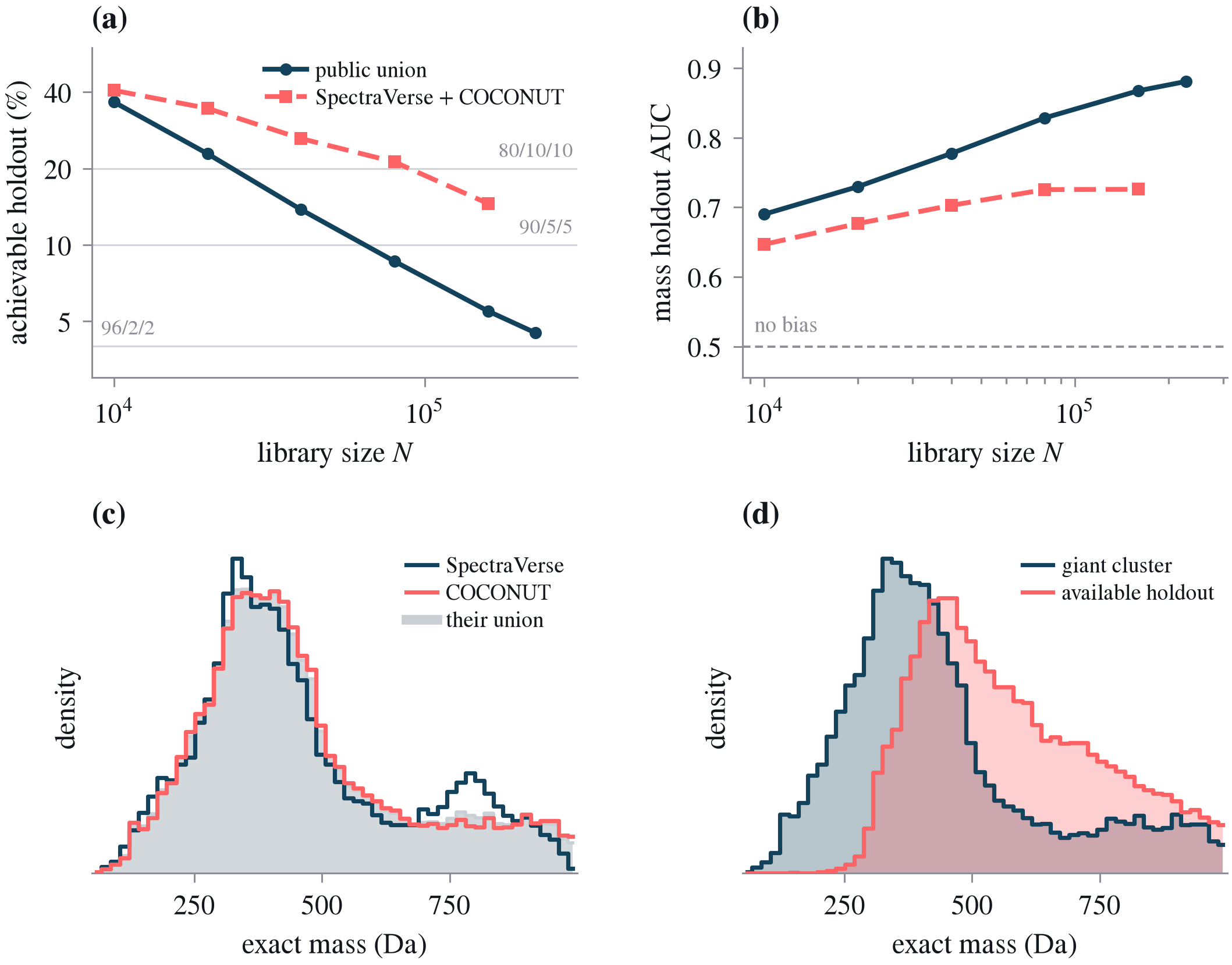}
\caption{\textbf{A natural-product library behaves the same way.} \textbf{(a)} Achievable holdout falls monotonically, $40.6\%$ to $14.6\%$. \textbf{(b)} Mass holdout AUC rises monotonically, $0.647$ to $0.726$, \textbf{(c)} Its two sources against their union. SpectraVerse and COCONUT
overlap closely, medians $406.3$ and $412.3$\,Da, which follows from both covering natural
products and small metabolites. \textbf{(d)} The split at $N = 160{,}000$: the giant sits at median $393.1$\,Da and everything outside it at $538.4$\,Da.}
\label{fig:coconut}
\end{figure}

COCONUT \cite{coconut} is an open collection of natural products. SpectraVerse is itself enriched in natural products and small metabolites, a consequence of the molecules of interest for MS/MS, so the two cover broadly similar chemistry, and their mass distributions nearly coincide (Figure~\ref{fig:coconut}). 

The library is built by substituting the 183,450 structures contributed by Enveda-180 with a matched number of compounds randomly sampled from COCONUT. This allows us to apply the same rung grid, with random subsampling, thresholding at $T=10$.

Both trends reappear (Figure~\ref{fig:coconut}a,b). The achievable holdout falls monotonically from 40.6\% to 14.6\%, and the mass holdout AUC rises monotonically from 0.647 to 0.726. At $N=160,000$ the giant sits at median 393.1 Da against 538.4 Da outside it, a 145 Da shift. It percolates more slowly as natural products are structurally more diverse than one deposit of drug-like synthetics, and at $N=160,000$ it retains nearly three times the union's holdout. 

Two caveats apply. Spectra exist for only 19.4\% of this library, and it is here as a counterfactual rather than an MS/MS benchmark anyone could build today. The ladder stops at $N=160,000$ due to compute constraints: the library percolates less and therefore skips less, dispatching 10.55\% of its candidates at that size against the union's 2.23\%, which puts a complete run several days of wall clock beyond the others.

\end{document}